\documentclass[10pt,twocolumn,letterpaper]{article}

\usepackage{graphicx}
\usepackage{amsmath}
\usepackage{amssymb}
\usepackage{booktabs}
\usepackage{svg}
\usepackage{multirow}

\usepackage[pagebackref,breaklinks,colorlinks]{hyperref}

\usepackage[capitalize]{cleveref}
\crefname{section}{Sec.}{Secs.}
\Crefname{section}{Section}{Sections}
\Crefname{table}{Table}{Tables}
\crefname{table}{Tab.}{Tabs.}

\begin{document}

\title{Comparison of Segmentation Methods in Remote Sensing for Land Use Land Cover}

\author{
Naman Srivastava, Joel D Joy,Yash Dixit, Swarup E, Rakshit Ramesh  \\
\textbf{Center of Data for Public Good} \\
\textbf{Indian Institute of Science, Bengaluru}
}
\date{}
\maketitle

\begin{abstract}
Land Use Land Cover (LULC) mapping is essential for urban and resource planning, and is one of the key elements in developing smart and sustainable cities.This study evaluates advanced LULC mapping techniques, focusing on Look-Up Table (LUT)-based Atmospheric Correction applied to Cartosat Multispectral (MX) sensor images, followed by supervised and semi-supervised learning models for LULC prediction. We explore DeeplabV3+ and Cross-Pseudo Supervision (CPS). The CPS model is further refined with dynamic weighting, enhancing pseudo-label reliability during training. This comprehensive approach analyses the accuracy and utility of LULC mapping techniques for various urban planning applications. A case study of Hyderabad, India, illustrates significant land use changes due to rapid urbanization. By analyzing Cartosat MX images over time, we highlight shifts such as urban sprawl, shrinking green spaces, and expanding industrial areas. This demonstrates the practical utility of these techniques for urban planners and policymakers. 
\end{abstract}

\section{Introduction}
\label{sec:intro}
High-resolution satellite multispectral imagery has emerged across numerous sectors, including defence, resource management, environmental analysis, and urban planning in the last two decades. These images for their fine spatial resolution cover and large sensor Field-of-View (FoV) covering extensive areas are sought as a great alternative for understanding and managing Earth's surface instead of field-based sampling (\cite{leprince2008monitoring}). However, manual annotation of such images for specific tasks is time-consuming and labour-intensive. To address this challenge, deep learning techniques, such as object detection and image segmentation, are used to automate and enhance the analysis of geospatial data. Among these tasks, Land Use Land Cover (LULC) mapping has gained significant importance, enabling governments and organizations to monitor resources, assess environmental changes, and develop informed policies. Deep learning algorithms can enhance the accuracy and efficiency of LULC mapping, contributing to sustainable development and resource management. However, there are a certain issues that need to be handled before using multispectral images with deep-learning applications. Raw multispectral images, typically captured as Top-of-Atmosphere (TOA) Digital Number (DN) values are influenced by atmospheric effects, such as scattering and absorption caused by aerosols, water vapour, and ozone, which contaminate the signal originating from the Bottom-of-Atmosphere (BOA) surface \cite{Singh2014}. For accurate surface characterization in deep learning applications, correcting these atmospheric interferences is essential. Therefore, converting TOA DN values to BOA reflectance using atmospheric correction (AC) is crucial to ensure analysis-ready data (ARD) and improve prediction accuracy\cite{dixit2024}.\\
Among deep learning models for Land Use and Land Cover (LULC) prediction, semantic segmentation plays a crucial role, with architectures like U-Net and DeepLab v3+ widely used. U-Net, initially designed for biomedical image segmentation \cite{ronneberger2015u}, features a U-shaped architecture with an encoder-decoder structure and residual connections. This design is highly effective for tasks that require precise feature localization \cite{jiwani2021semantic, robinson2022fast, sirko2021continental}. In contrast, DeepLab v3+, developed by Google DeepMind, leverages atrous convolution to expand the field of view, capturing multi-scale contextual information. This makes it particularly adept at identifying various land types such as urban areas, forests, water bodies, and agricultural lands \cite{chen2018encoder, fan2022land, jiwani2021semantic}.
Semi-supervised segmentation has been explored a lot in Remote Sensing including techniques such as consistency regularisation and iterative training \cite{wang2022self,zhang2020semi,wang2022semi,sanchez2024self,li2023one,zhu2023survey}. Another popular approach in semi supervised learning is Cross Pseudo Supervision (CPS) which improves performance by enforcing consistency  between two networks through cross supervision \cite{chen2021semi}. However, CPS faces challenges like overfitting and class imbalance. While several studies have applied semi-supervised techniques for LULC prediction, addressing issues like class imbalance and geographic contexts (e.g., coastal areas, farmlands) \cite{lu2022simple,cenggoro2017classification,xu2021semantic,sertel2022land,wang2021mask}, robust LULC mapping models must account for variability in satellite imagery caused by time of day, season, and regional differences in building and vegetation appearance across India’s diverse climatic and environmental conditions. A generalized model should perform well across terrains such as coasts, plateaus, farmlands, river basins, and metropolitan areas.
The Dual-debiased Heterogeneous Co-training (DHC) framework introduces adaptive loss-weighting strategies: Distribution-aware Debiased Weighting (DistDW) and Difficulty-aware Debiased Weighting (DiffDW) \cite{wang2023dhc}. DistDW adjusts class weights based on batch distribution, while DiffDW dynamically weights classes based on changes in dice scores. These strategies effectively address biases and class imbalance. Further improvement by Wang et al. \cite{wang2024Genssl} integrates these strategies into a generalized semi-supervised framework, addressing class imbalance, Unsupervised Domain Adaptation, and domain generalization.\\ In most semi-supervised learning approaches, distinct sets of labeled and unlabeled images are used. Although numerous accurately and densely labeled remote sensing datasets exist \cite{codegoni2023tinycd,ISPRS_Potsdam,castillo2022semi}, they vary in satellite image resolution. Users often need to train their Land Use and Land Cover (LULC) models on images with different resolutions from those in benchmark datasets, which can lead to suboptimal performance. Preparing ground truth labels for such cases often requires manual annotation or leveraging open data sources like OpenStreetMap\cite{OpenStreetMap}. However, these sources can suffer from sparse labeling, misclassification, and outdated information due to ongoing construction and deconstruction. Studies have applied weakly supervised segmentation techniques for Land Use Land Cover \cite{dixit2024,schmitt2020weakly,nivaggioli2019weakly,wang2023sdcdnet}. In this study, we apply Atmospheric correction techniques followed by  popular supervised and semi-supervised learning techniques and evaluate their performance on a sparsely labeled dataset.

\begin{figure*}[h]
    \centering    \includegraphics[width=1.0\textwidth ]{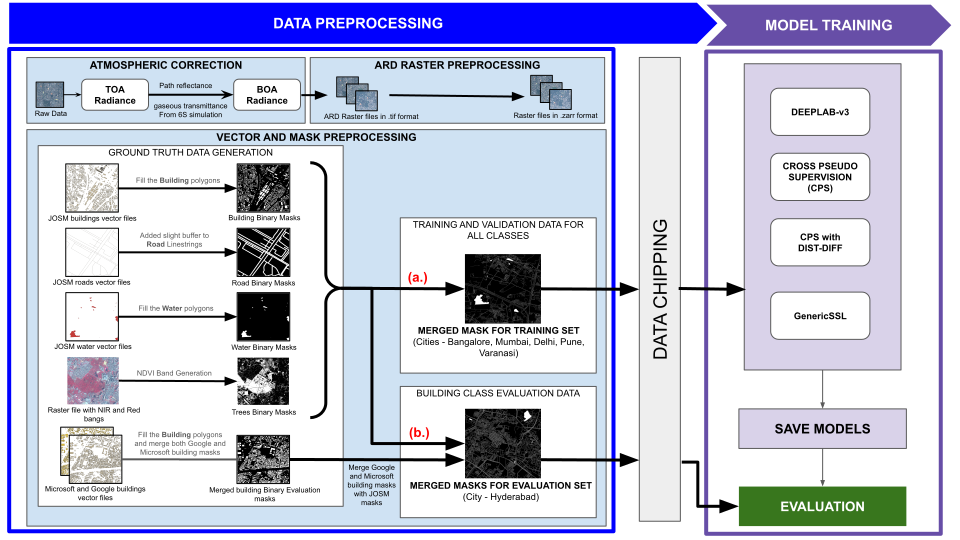}
    \caption{Workflow for LULC Segmentation. \textcolor{red}{\textbf{(a.)}}:  The Merged Training masks are created by combining the binary class masks of the concerned cities which have been generated by using the JOSM vector files. Apart from these classes, the remaining areas are classified as the "Other" class. During training, we focus on a subset of data that is densely populated with classes and extract patches from it.
    \textcolor{red}{\textbf{(b.)}}: The Merged Evaluation masks are created by utilizing the output binary masks of roads, water, trees, combined data of buildings (JOSM, Microsoft and Google), and treating the remaining areas as the "Other" class. During model evaluation, data is chipped without imposing class heavy constraints.}
    \label{fig:pipeline}
\end{figure*}

\section{Methodology}The proposed model architecture involves several key stages: preparing raster masks from vector data, dividing large satellite images into smaller patches for model training, and training the model to predict segmentation outputs. Finally, the model's performance is evaluated to assess its effectiveness in predicting land use and land cover class masks. Figure \ref{fig:pipeline} illustrates the sequence of these stages in the machine learning operations pipeline.
\subsection{Data Sources}

\subsubsection{Raster Data}
This study used high-resolution multispectral imagery (approximately $1.134 \, \text{m}^2/\text{px}$) covering the Blue (0.45-0.52 $\mu$m), Green (0.52-0.59 $\mu$m), Red (0.62-0.68 $\mu$m), and NIR (0.77-0.86 $\mu$m) spectral bands. These scenes were acquired over six Indian cities by the Multi-spectral (MX) sensor on board the CartoSAT-3 satellite, operated by the Indian Space Research Organisation (ISRO). To ensure sufficient heterogeneity in land use and land cover (LULC) classes, each tile was cropped into smaller subsets as part of the preprocessing workflow (see Fig.~\ref{fig:pipeline}). Each scene, covering a specific area, had less than 20\% cloud coverage but was affected by sensor noise and atmoshperic interactions, mainly Rayleigh and Aerosol scattering and absorption from gases (details in section \ref{ssec:ARD}). The individual bands were consolidated into raster stack in Zarr format \cite{zarr} instead of Xarrary \cite{xarray} for its ability to save and load chunks separately, optimizing memory usage.

\subsubsection{Vector Data}
The vector data for training and evaluation were sourced as follows:\\
\textbf{Training}: Vectors for Bangalore, Mumbai, Pune, Varanasi, and Delhi were obtained from OpenStreetMap (OSM)\cite{OpenStreetMap} using the JavaOpenStreetMap (JOSM) editor. These were saved as GeoJSON files within the geospatial bounds of corresponding raster files.\\
\textbf{Evaluation:} Evaluation data included vector datasets of predicted buildings from Microsoft, Google, and JOSM, filtered to match the Area of Interest (AOI) in Hyderabad. The vectors were aligned with ISRO raster images and saved as GeoJSON files after removing redundant data.

\subsection{Data Preparation}
\begin{figure}[h]
    \centering
    \includegraphics[width=0.4\textwidth]{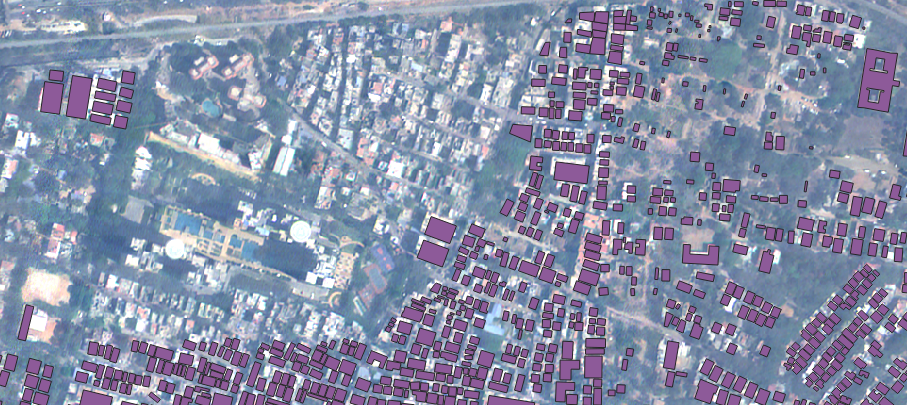}
    \caption{Sparse Label (Buildings)}
    \label{fig:sparse}
\end{figure}

The data preparation process involved generating class-specific binary masks and subsequently merging them into a comprehensive multi-class mask. Initially, vector data and corresponding raster files were aligned using a shared Coordinate Reference System (CRS). Various vector geometries were utilized: Linestrings for roads and Polygons/Multipolygons for buildings and water bodies, each processed differently to generate the class-specific ground truth mask. For roads, a 3-pixel buffer was applied to the Linestrings using the Shapely library \cite{shapely2007}, while buildings, represented by Polygons, were fully filled (Fig. \ref{fig:sparse}).  Few studies have used a building buffer techniques \cite{robinson2022fast}, where building borders were treated as a separate class. this approach may increase model complexity and hence not included in the study. Vegetation ground truth masks were generated using the Normalized Difference Vegetation Index (NDVI), with manual inspection determining the threshold for binary classification. The binary masks were stored in Zarr format after mapping them to their corresponding MX images.\\

To accurately represent different land use classes, we had to resolve overlaps such as buildings overlapping vegetation or roads crossing water bodies. To handle these conflicts, we combined the binary masks for buildings, roads, and water bodies into a single multi-class mask. Each class was assigned a unique integer value, where more dominant classes (like buildings) were given lower values. In the event of overlaps, the maximum value across the individual masks was chosen, ensuring that the minority class was accurately reflected. Areas with absent labels were classified as "Other".
Finally, the dataset was divided into chunks of shape (4 x 256 x 256) to generate image patches for training. patches with more than 50\% missing values on the multi-spectral image, or more than 65\% pixels belonging to the "Other" class were discarded, enhancing the quality of the training dataset. This filter ensured that the training data has more than 35\% labelled area. Using this method, we obtain 60848 image patches, covering a total land area of 5091.94 $km^2$. nevertheless, this filtering strategy does not guarantee 100\% labelled data for training as seen in Figure \ref{fig:sparse}, and does not prevent potential mislabeling in the ground truth masks. 
\subsection{DN to ARD coversion}
\label{ssec:ARD}
The Top-of-Atmosphere (TOA) apparent reflectance ($\rho^*$) can be related to the Bottom-of-Atmosphere (BOA) reflectance, assuming a Lambertian surface, as described in \cite{6s}

\begin{align}
\label{eq:rtm_atmos}
\rho^*\left(\theta_s, \theta_v, \Delta \phi, \lambda\right)= & \ T_g\left(\theta_s, \theta_v, \lambda\right) \times \bigg[\rho_{r a}\left(\theta_s, \theta_v, \Delta \phi, \lambda\right) \nonumber\\
& + \frac{T\left(\theta_s, \lambda\right) T\left(\theta_v, \lambda\right) \rho_s(\lambda)}{1-S(\lambda) \rho_s(\lambda)}\bigg]
\end{align}
Where, $\theta_s$, $\theta_v$, $\Delta \phi$ and $\lambda$ are the sun zenith angle, viewing zenith angle, relative azimuth angle and wavelength, respectively. $\rho$ denotes apparent reflectance obtained as $\rho = \frac{\pi L}{F_0 \mu_s}$ (See \cite{mukherjee2024} for details). $\rho_s$ represents BOA (surface)reflectance and $\rho_{ra}$ is the path reflectance from the combined interaction of aerosols and molecules. $T\left(\theta_s, \lambda\right)$ and $T\left(\theta_v, \lambda\right)$ are the upward and downward atmospheric transmittance, $S$ is the spherical albedo and $T_g$ refers to gaseous (Ozone and Water Vapour) transmittance.

6S \cite{6s} is used to simulate Look-Up Tables (LUTs) storing a large set of AC coefficients ($a,b$, obtained by rearraging the unknown components in Eq. \ref{eq:rtm_atmos} apart from the $\rho_s$). The coefficients $a$ represents $\rho_{ra}$, a function of aerosol optical thickness and atmospheric molecular Rayleigh scattering, whereas, $b$ expresses gaseous absorption and transmittance, essentially function(s) of atmospheric water vapour and ozone, primarily along with the target surface elevation and acquisition geometry. The LUT is stored as 5-D interpolatable object to interpolate approximated values of $a$ and $b$ for a given gaseous concentration, (Water and ozone) aerosol optical thickness (AOT), surface elevation and viewing geometry ($\theta_s, \theta_v$) obtained as mean over the extent for each band present in MX scenes.

\subsection{Model Training}
Our comparative study considers both Supervised and Semi-Supervised techniques. DeepLabV3+ is used as a baseline model architectures for supervised learning. For Semi-supervised methods, we incorporate Cross Pseudo supervision and later on combine it with two de-biased heterogeneous model training approaches:  distribution-difficulty aware and difficulty-difficulty aware modeling.
\subsubsection{Supervised Learning}
We implemented the DeeplabV3+ model with the EfficientNet backbone from the Segmentation Models Pytorch library \cite{Iakubovskii:2019} as a baseline for our study. To address class imbalance, the model was trained using a weighted pixel-wise cross-entropy loss, with class weights set inversely proportional to class abundance.
\subsubsection{Cross Pseudo Supervision}

In the cross-pseudo supervision method, two DeepLabv3+ models with EfficientNet backbones, $f(\theta1)$ and $f(\theta2)$, are initialized differently and trained on sparsely labeled data, allowing both supervised and unsupervised loss computation using the same batch of labeled data, unlike the approach by Xiaokang Chen et al. (2021) \cite{chen2021semi}. The supervised loss consists of the Hausdorff Erosion loss \cite{HausdorffER}, $L^{HF}_{Sup}$, weighted by class-specific weights $\alpha_{\text{w}}$, and the Weighted Cross Entropy Loss, $L^{WCE}_{Sup}$. Both losses are normalized by a constant $K$ based on the number of labeled images and image dimensions. The total supervised loss, $L_{Supervised}$, is the sum of $L^{HF}_{Sup}$ and 0.5 times $L^{WCE}_{Sup}$, as shown in Equation \ref{eq:sup}.

\begin{equation}
    \begin{aligned}
        K &= {|D_{\text{SL}}|} \times {W\times H}
    \end{aligned}
    \label{eq:constant}
\end{equation}

\begin{equation}
    \begin{aligned}
        L^{*}_{Sup} &= \frac{1}{K} \times \sum_{X\in D_{\text{SL}}} \sum_{w=1}^{5} \sum_{i=0}^{W\times H}  \alpha_w \times \bigl( l_{\text{hd}}(P_{1iw}, y^{*}_{iw})  \\
        & \hspace{4cm} + l_{\text{hd}}(P_{2iw}, y^{*}_{iw}) \bigr)
    \end{aligned}
    \label{eq:Haursd}
\end{equation}

\begin{equation}
    L_{Supervised} = L^{HF}_{Sup} + 0.5* L^{WCE}_{Sup}
    \label{eq:sup}
\end{equation}

To calculate the pseudo supervised loss in CPS, we use weighted cross-entropy between the softmax outputs of model 2 and the one-hot encoded outputs from model 1, and vice versa. This is detailed in Equation \ref{eq:wce_cps}, with the total pseudo supervised loss represented as $L^{WCE}_{CPS}$.

\begin{equation}
    \begin{aligned}
        L^{WCE}_{CPS} &= \frac{1}{K} \times \sum_{X\in D_{\text{SL}}} \sum_{w=1}^{5} \sum_{i=0}^{W\times H}  \alpha_w \times \bigl( l_{\text{ce}}(P_{1iw}, y_{2iw}) \\ 
        & \hspace{4cm} +  l_{\text{ce}}(P_{2iw}, y_{1iw}) \bigr)
    \end{aligned}
    \label{eq:wce_cps}
\end{equation}
Here, $P_{1iw}$ and $P_{2iw}$ denote the logit outputs from the two models, $y^{*}_{iw}$ are the ground truth masks, and $y_{1iw}$ and $y_{2iw}$ are the one-hot vectors derived from $P_{1iw}$ and $P_{2iw}$, respectively. The total loss combines the supervised loss, $L_{\text{Supervised}}$, with the cross-pseudo supervision loss, $L^{\text{WCE}}_{\text{CPS}}$, as shown in Equation \ref{eq:totalLoss}. A sigmoid ramp-up function, detailed by Laine et al. (2016) \cite{laine2016temporal}, gradually increases the weight of $L^{\text{WCE}}_{\text{CPS}}$ from 0 to 0.1 over a specified number of epochs, adjusting the trade-off parameter $\lambda$ based on the total epochs \( T \), current epoch \( t \), and ramp-up length \( r \).

\begin{equation}
    L_{Total} = L_{Supervised} + \lambda \cdot L^{WCE}_{CPS}
    \label{eq:totalLoss}
\end{equation}

\begin{equation}
    \text{rampup}(r, t, T)=
\begin{cases} 
0 & \text{if } t = 0, \\
0.1 \cdot \exp\left(-5.0 \left(1.0 - \frac{t}{T}\right)^2\right) & \text{if } 1 \leq t \leq r, \\
0.1 & \text{otherwise}.
\end{cases}
\label{eq:rampup}
\end{equation}

Here, the trade-off parameter $\lambda$ is adjusted according to the ramp-up function, where \( T \) represents the total number of epochs, \( t \) denotes the current epoch, and \( r \) specifies the ramp-up length.

\subsubsection{ Distribution - Difficulty Aware Framework for Sparsely labeled Geospatial Images}
The study modifies the Cross Pseudo Supervision (CPS) approach by integrating dynamic weighting strategies from the DHC framework \cite{wang2023dhc}. Unlike DHC, which uses both labeled and unlabeled data, this operates on a sparsely labeled dataset, computing supervised and unsupervised losses on the same image patches. It independently applies difficulty-aware and distribution-aware weighting strategies, with a sigmoid ramp-up strategy gradually increasing the impact of the unsupervised loss.

\textbf{Distribution-Aware Weight Calculation (Dist)}: The distribution-aware weighting strategy tackles class imbalance by dynamically assigning class-specific weights in both the supervised and CPS loss functions. This approach calculates the pixel count for each class from the one-hot encoded outputs and computes the ratio of pixels in the most abundant class to those in each specific class, denoted as \( R_k \) (Equation \ref{eq:probs}), where \( K \) is the total number of classes and \( k \) ranges from 1 to \( K \).

\begin{equation}
    R_k = \frac{\max{\left \{ N_k \right \}_{i=1}^{K}}}{N_k}
    \label{eq:probs}
\end{equation}

Subsequently, the logarithmic ratio of \( R_k \) for each class \( k \) to the maximum logarithmic value of \( R_k \) across all classes is used to calculate the weights \( w_k \) (Equation \ref{distW}).

\begin{equation}
    w_k = \frac{\log(R_k)}{\max{\left \{ \log(R_k) \right \}_{i=1}^{K}}} 
    \label{distW}
\end{equation}

Initial weights are computed over the entire training dataset, with subsequent recalculations for each batch. An Exponential Moving Average (EMA) of class weights is applied to ensure continuous updates during training, as shown in Equation \ref{eq:distEMA}, where \( \beta \) denotes the momentum and \( W_t \) represents the weights at epoch \( t \), both of which are tunable hyperparameters.

\begin{equation}
    \begin{aligned}
    W_{t}^{dist} &\leftarrow \beta W_{t-1}^{dist} + (1-\beta) W_{t}^{dist} \\
    & \hspace{0.2cm} , W_{t}^{dist} = [w_1,w_2...w_k]
    \end{aligned}
    \label{eq:distEMA}
\end{equation}

\textbf{Difficulty-Aware Weight Calculation (Diff)}: The difficulty-aware strategy uses the Soft Dice coefficient to compute dynamic weights for each class, assessing the similarity between logits and ground truth masks. Initially, weights are set uniformly for all classes. The Difficulty-Aware model calculates the Dice coefficient for each class and evaluates changes over epochs. Positive changes indicate learnability, while non-positive changes suggest unlearnability, as defined by Equations \ref{eq:unlearn} and \ref{eq:learn}. The difficulty-adjusted weight for each class is computed as
\begin{equation}
    du_{k,t} = \sum_{t - \tau}^{t} \mathbb{I}(\Delta \leq 0) \ln \left( \frac{\lambda_{k,t}}{\lambda_{k,t-1}} \right),
    \label{eq:unlearn}
\end{equation}

\begin{equation}
    dl_{k,t} = \sum_{t - \tau}^{t} \mathbb{I}(\Delta > 0) \ln \left( \frac{\lambda_{k,t}}{\lambda_{k,t-1}} \right),
    \label{eq:learn}
\end{equation}

\begin{equation}
    d_{k,t} = \frac{du_{k,t}+\varepsilon }{dl_{k,t}+\varepsilon }
    \label{eq:diffW}
\end{equation}

\begin{equation}
    w_{k}^{diff} = w_{\lambda_{k,t}}.(d_{k,t})^{\alpha}
    \label{eq:diffWWQ}
\end{equation}

Here, \( du_{k,t} \) and \( dl_{k,t} \) represent the unlearned and learned metrics, respectively, with \( \varepsilon \) as a small constant to prevent division by zero. The final class weights are normalized and scaled, prioritizing classes that are harder to learn by adjusting to the learning progress.
\subsubsection{Generic Self Supervised Learning (SSL)}
We implement a self-supervised learning (SSL) framework for representation learning derived from Aggregating and Decoupling framework \cite{wang2024Genssl}. The implemented model consists of a single encoder, followed by three decoders namely Distribution aware (Dist), Difficulty aware (Diff) and Ensemble (Ens) decoder. The losses for Dist and Diff decoders are computed against the ground truth labels, whereas the loss for Ens decoder is computed against the ensemble of predictions obtained by the other two decoders. While \cite{wang2024Genssl} focuses on training two decoders using labelled data and the third decoder using the only the unlabelled data, this study trains all three decoders using sparsely labeled, multi-spectral remote sensing imagery.
\subsection{Post Processing}
To mitigate inaccuracies from dividing large geospatial images into smaller patches for model training and prediction, we implemented two post-processing techniques. Initially, \textbf{Prediction Ensembling} was used, where predictions from the Distribution Aware and Difficulty Aware models were combined by averaging the softmax probabilities of each class. Consecutively, \textbf{Prediction Merging} was applied by generating overlapping patches using a sliding window on the evaluation dataset. The predictions from these patches were merged back into the full image using a max pooling approach, ensuring that predictions accounted for the surrounding context, thus improving accuracy and overall evaluation scores. To compute test scores, softmax probabilities were converted to binary outputs using Equation \ref{eq:softmax}, where $\rho_{i,j}$ and $x_{i,j}$ represent the binary value and softmax value for the $(i,j)^{th}$ pixel, respectively.

 \begin{equation}
         \rho(i,j)= 
\begin{cases}
    1,& \text{if } x_{i,j}\geq \textit{threshold}\\
    0,              & \text{otherwise}
\end{cases}
\label{eq:softmax}
 \end{equation}

\section{Results and Discussion}
\begin{figure}[h]
    \centering
\includegraphics[width=0.45\textwidth]{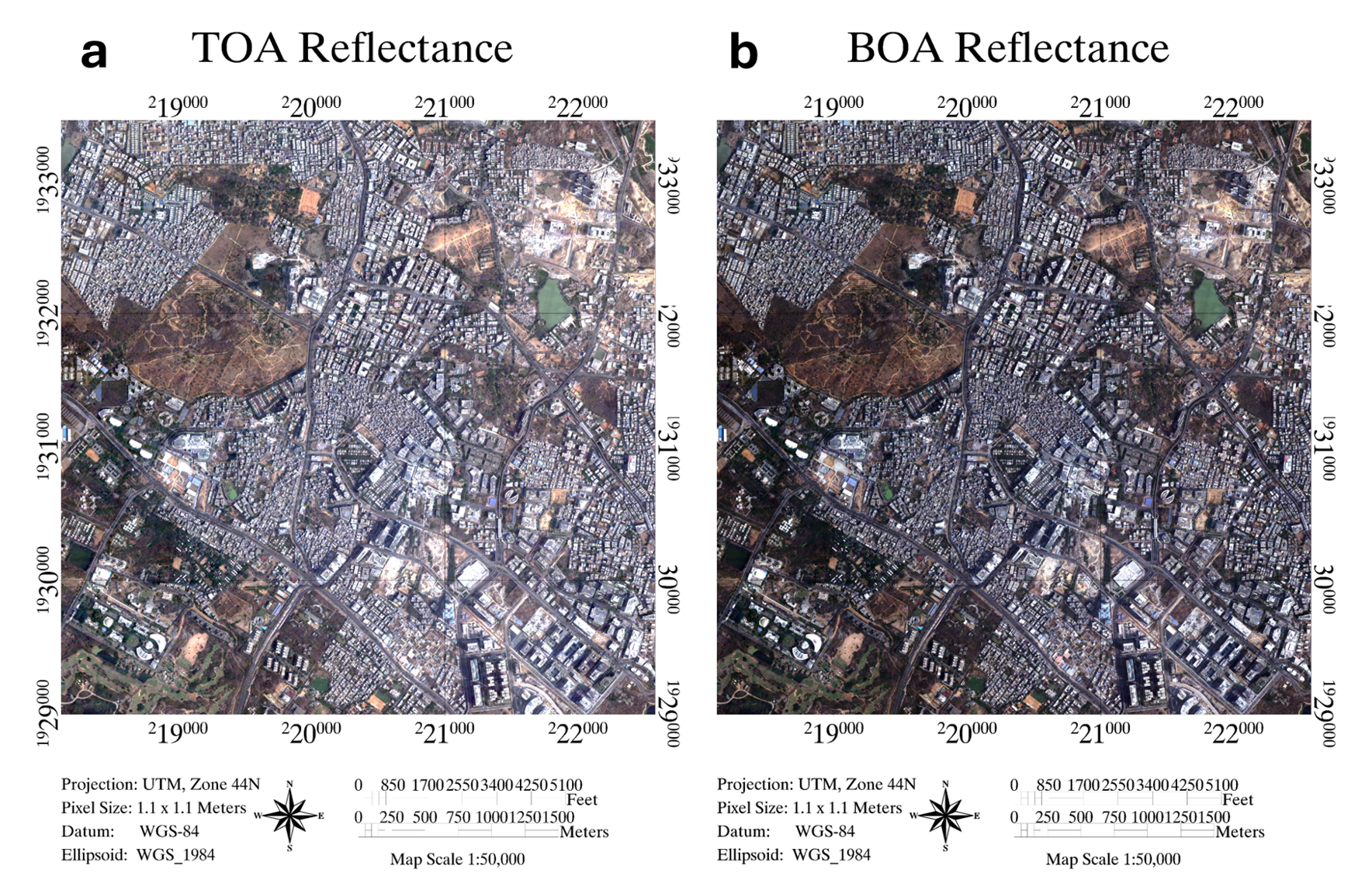}   
   
    \caption{(a,b) shows sensor acquired TOA reflectance and ARD obtained BOA reflectance, respectively}
    \label{fig:evidence}
\end{figure}

\begin{figure}[h]
    \centering    \includegraphics[width=0.45\textwidth]{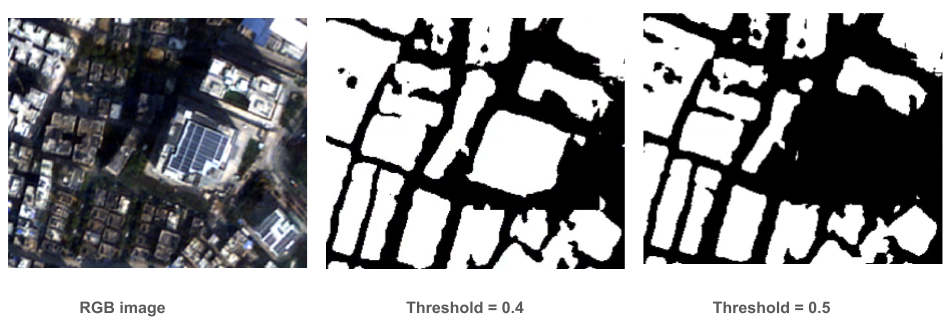}  
    \centering    \includegraphics[width=0.45\textwidth]{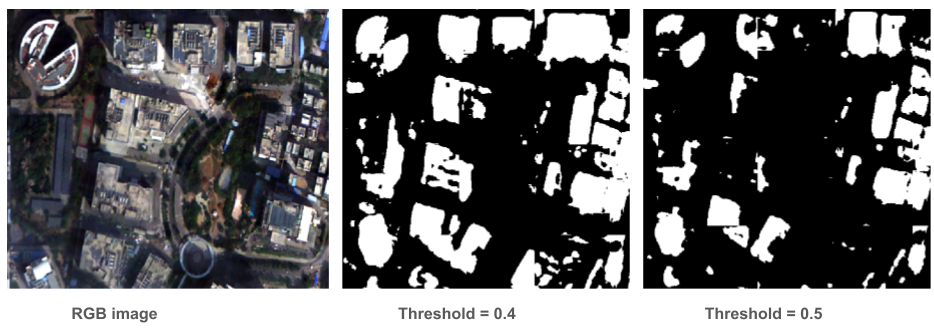}
    \caption{Mask Threshold comparison for a small patch of the image.}
    \label{fig:evidence1}
\end{figure}

The LUT-based ARD pipeline was applied to a patch of MX image acquired over the Indian city of Hyderabad on March 12, 2023. The corrected image (Fig. \ref{fig:evidence}b) demonstrated a clear improvement, effectively eliminating atmospheric distortions that often result in a hazy appearance in the uncorrected $\rho^{}$ (Fig. \ref{fig:evidence}a). Supervised and unsupervised learning models as listed in \ref{tab:recall-scores} were evaluated by calculating Recall and MIoU scores, detailed in Table \ref{tab:recall-scores}, for the patch representing $24.6 \text{ km}^2$ area of the Hyderabad Cartosat image. The scores are calculated on a binary outputs obtained using \textit{0.4} as a threshold value in Equation \ref{eq:softmax}. Figure \ref{fig:evidence1} depicts the difference in prediction mask obtained by using \textit{0.4} as threshold instead of the standard \textit{0.5} 
\begin{table*}[ht]
    \centering
    \caption{Recall and Mean Intersection-over-Union Scores For Supervised and Semi-Supervised Model}
    \begin{tabular}{|p{6.5cm}|c|c|c|c|c|c|}
    \hline
        \textbf{Model/Framework} & \textbf{Metric} & \textbf{Buildings} & \textbf{Water} & \textbf{Trees} & \textbf{Roads} & \textbf{Average} \\
        \hline
        \multirow{2}{*}{CPS with Dist-Diff} & \textbf{Recall} & {75.01} & {73.03} & \textbf{96.61} & \textbf{62.91} & 76.89 \\
          & MIoU & \textbf{42.01}  & \textbf{49.65} & 56.59 & 20.32 & \textbf{42.14} \\
         \hline
         \multirow{2}{*}{CPS} & Recall & \textbf{75.09} & \textbf{86.96} & 93.41 & 62.89 &\textbf{79.59}\\
          & MIoU & 40.34 & 10.70 & \textbf{64.12} & \textbf{20.66} & 33.95 \\
         \hline
         \multirow{2}{*}{GenSSL with Ens} & Recall& 39.91 & 66.84 & 86.34 & 39.40 & 58.12 \\
          & MIoU & 29.75 & 42.70 & 69.28 & 17.76 & 39.87  \\
         \hline
         \multirow{2}{*}{GenSSL with Ens-Diff} & Recall& 43.41 & 72.44 & 86.19 & 40.17 & 60.55 \\
          & MIoU & 27.37 & 6.70 & 69.28 & 17.58 & 30.23  \\
         \hline
         \multirow{2}{*}{GenSSL with Ens-Dist} & Recall& 49.47 & 84.61 & 92.29 & 55.11 & 70.37 \\
          & MIoU & 31.06 & 4.13 & 50.30 & 17.95 & 25.86  \\
         \hline
         \multirow{2}{*}{GenSSL with Ens-Dist-Diff} & Recall& 54.96 & 89.61 & 84.77 & 53.72 & 70.79\\
          & MIoU & 23.43 & 2.37 & 47.53 & 17.01 & 22.59 \\
         \hline
         \multirow{2}{*}{Distribution Aware (No Post Processing)} & Recall & 54.08 & 68.69 & 93.74 & 44.99 & 65.38\\
          & MIoU & 36.71 & 50.14 & 58,07 & 19.90 & 40.96 \\
         \hline
         \multirow{2}{*}{Difficulty Aware (No Post Processing)} & Recall & 38.92 & 63.88 & 79.80 & 36.65 & 54.81\\
          & MIoU & 29.73 & 51.36 & 66.84 & 19.27 & 41.80 \\
      
         \hline
         \multirow{2}{*}{Deeplabv3+} & Recall & 59.90 & 57.44 & 51.72 & 41.00 & 52.52 \\
          & MIoU & 36.33 & 41.18 & 44.35 & 17.99 & 34.96 \\
         \hline
    \end{tabular}
    \label{tab:recall-scores}
\end{table*}
Evaluating the performance of our model using standard metrics like Accuracy or Dice score is challenging due to the sparse labelling of the ground truth images. These metrics would not accurately assess the model's performance as there are many instances where the ground truth labels are incorrectly annotated with respect to the image patch in consideration. Instead, we use a metric that emphasizes on True Positives and False Negatives. Therefore, we find Recall, to be a more suitable metric for our evaluation. 
\begin{figure*}[btp]
    \centering    \includegraphics[width=0.95\textwidth ]{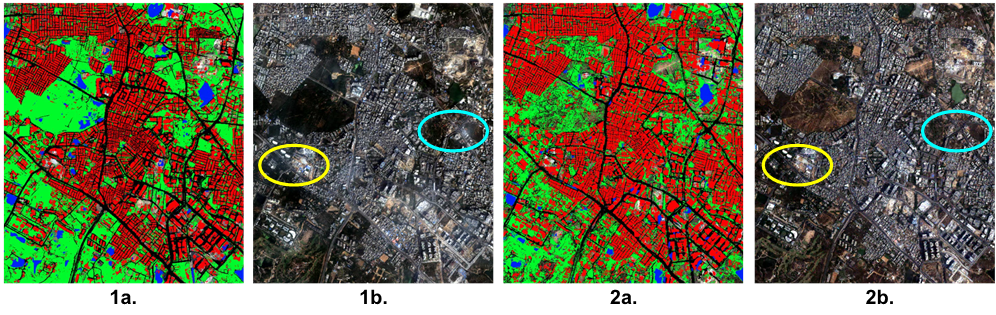}
    \caption{Images with indexes starting with \textbf{"1"} correspond to the Hyderabad image that were captured in December 2020, and the ones starting with \textbf{"2"} i correspond to the image captured in March 2023. Images with indexes ending with \textbf{"a"} indicate the predicted masks, and the ones ending with \textbf{"b"} indicate the corresponding original NRGB images captured from CartoSAT Satellite. In \textbf{"1a."} and \textbf{"2a."}, the Red Colour in the predicted masks generated by the proposed model within this study represents buildings, Green colour represents Vegetation, Black colour represents Roads, and Blue represents Water Bodies. In \textbf{"1b"} and \textbf{"2b"}, the cyan circle marks the Hitech City area, and yellow circle marks the Gachibowli area.  }
    \label{fig:Hyderabad-images}
\end{figure*}
Table \ref{tab:recall-scores} shows that the Cross Pseudo Supervision (CPS) approach significantly outperforms supervised models like DeepLabV3+. CPS improves performance by addressing class imbalance with class weights inversely proportional to class abundance. We find that on integrating the dynamic weighting strategy Dist-Diff in the CPS model, the average of recall scores of classes see a decrease but the Mean Intersection over Union (MIoU) scores are improved. boosting recall scores for abundant classes like trees while improving MIoU performance on minority classes such as water. Compared to \cite{robinson2022fast}, which used very high-resolution images (50cm) and evaluated on sparsely populated regions, our model, trained on high resolution images (1.14m) and evaluated on densely populated areas, shows strong performance. It is worth noting that even though the scores of Generic SSL are not at par with CPS, it was found to be computationally less expensive, requiring shorter time and lesser space to train and deploy, as compared to the other semi supervised models implemented in this study, while still outperforming supervised learning model Deeplab. We also observe that the post processing techniques including prediction merging and ensembling improve the recall scores. 
\begin{table*}[h]
    \centering
    \caption{Classwise change in Land-Use Land-Cover in Hyderabad}
    \begin{tabular}{|c|c|c|c|c|}
    \hline
         \textbf{Classes} & \textbf{Hyd 2020 class area} ($km^2$)  & \textbf{Hyd 2023 class area} ($km^2$) & \textbf{Area Gained/Lost} ($km^2$) & \% \textbf{Change} \\
         \hline
         \hline
         Building & 8.2146  & 9.0611 & 0.8464 & 10.3046\% \\
         \hline
          Road & 11.2745 & 14.8226 & 3.5481 & 31.4704\% \\
         \hline
         Water & 0.8232 & 0.8023 & -0.0209 & -2.5389\% \\
         \hline
         Trees & 10.1444 & 7.0209 & -3.1235 & -30.7904\% \\
         \hline
    \end{tabular}
    
    \label{tab:area}
\end{table*}
\subsection{Case Study on Hyderabad}
This case study investigates the Land Use and Land Cover (LULC) changes using MX ARD within Hyderabad, with a particular focus on its rapidly growing IT hub. Areas such as Gachibowli and Hitech City, which have experienced extensive urbanization and economic growth, are of particular interest due to significant land acquisition by corporate entities and consequent tree felling. Understanding the LULC dynamics in these regions is crucial to assess the environmental impact of the IT sector's expansion over time.\\
The study analyzes equivalent areas in Hyderabad using MX images captured at different times, specifically in December 2020 and March 2023. The area of interest spans 24.3879 square kilometers, encompassing a key zone within the city that includes the major IT hub. The methodology described in the Post-Processing section was employed to predict and merge the image patches back to their original NRGB size. After merging, a thresholding mechanism, as outlined in the model evaluation process, was applied to the softmax outputs. The results are summarized in Table \ref{tab:recall-scores}.\\
The analysis reveals significant changes in land use, particularly in the northern part of Gachibowli and Hitech City, where substantial tree removal and the expansion of concrete structures have occurred (Table \ref{tab:area}). Additionally, central Gachibowli has seen the filling of a lake to make way for new buildings and roads (Figure \ref{fig:Hyderabad-images}).  
\section{Conclusion}
Our work makes a significant contribution to LULC (Land Use/Land Cover) analysis by enhancing the accuracy of classification through improved preprocessing and model evaluation. Atmospheric correction (AC) transformed raw multispectral images into Analysis Ready Data (ARD), boosting the performance of \textit{state-of-the-art} supervised and semi-supervised models. These models, evaluated on a patch in Hyderabad city using Recall and MIoU metrics, showed that CPS achieved the highest Recall, while CPS with Dynamic Weighting (DHC) excelled in MIoU. The improvement in Recall can be substantially linked to the AC process, as it provides true ground reflectance feature into training models increasing the reliability of LULC classification. Both visual and statistical evalution further identified significant LULC changes over three years for a hotspot in Hyderabad, i.e., reduction in vegetation caused by the rapid urbanization. Overall, this study showcases our AC integrated segmentation method’s effectiveness in capturing land use shifts that can have significant contributions to more precise environmental and urban planning. 

\appendix
\section{Appendix}
\subsection{Model Training Details}
We implemented the proposed frameworks mentioned in the methodology section with PyTorch, using 2 NVIDIA
4090 GPUs, 2 i9 13th gen CPUs and 256GB of RAM. We setup MLOps pipelines(the workflows of which have been briefed in Figure \ref{fig:pipeline}) for all of the mentioned frameworks in the Model Training Section using an open-source tool called Kedro. Argo was the workflow engine used in Katib workers. Katib was deployed and orchestrated by Kubeflow. Since, Argo supports Directed Acyclic Graphs (DAGs) and step-based workflows, it allowed us to run Kedro pipelines within Argo. This enabled multiple pipelines to run in parallel, which in turn reduced execution times. The network parameters were optimized using Adam with an initial learning rate of 0.001, and we employed a “poly” decay strategy following \cite{isensee2021nnu}. Given that the NRGB images which were used were essentially noisy, they were used as-is during the training stage to prevent over-fitting. The networks were trained for 350 epochs with a batch size of 4.

{\small
\bibliographystyle{ieee_fullname}
\bibliography{References}
}

\end{document}